\documentclass{article}

\usepackage{arxiv}

\usepackage[utf8]{inputenc} 
\usepackage[T1]{fontenc}    
\usepackage{hyperref}       
\usepackage{url}            
\usepackage{booktabs}       
\usepackage{amsfonts}       
\usepackage{nicefrac}       
\usepackage{microtype}      
\usepackage{lipsum}		
\usepackage{graphicx}
\usepackage{natbib}
\usepackage{doi}

\usepackage{array}
\usepackage[caption=false,font=normalsize,labelfont=sf,textfont=sf]{subfig}
\usepackage{textcomp}
\usepackage{stfloats}
\usepackage{url}
\usepackage{verbatim}
\usepackage{graphicx}
\def\BibTeX{{\rm B\kern-.05em{\sc i\kern-.025em b}\kern-.08em
    T\kern-.1667em\lower.7ex\hbox{E}\kern-.125emX}}
\usepackage{balance}
\usepackage{hyperref}
\usepackage{algorithm}
\usepackage{algpseudocode}
\usepackage{booktabs}
\usepackage{multirow,booktabs}
\usepackage{amsmath}
\usepackage[table]{xcolor}
\usepackage{xurl} 
\usepackage{subfig}
\usepackage[normalem]{ulem}
\usepackage{breqn}
\usepackage{comment}
\usepackage{caption}
\usepackage{amsmath}
\usepackage{amssymb}
\usepackage{algpseudocode}
\usepackage{array}
\newcolumntype{P}[1]{>{\centering\arraybackslash}p{#1}}
\usepackage{lineno}
\usepackage{cite}

\title{CMAViT: Integrating Climate, Management, and Remote Sensing Data for Crop Yield Estimation with Multimodel Vision Transformers}

\author{ \href{https://orcid.org/0000-0001-9718-7518}{\includegraphics[scale=0.06]{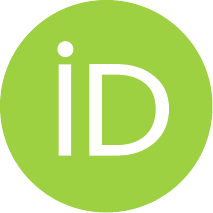}\hspace{1mm}Hamid Kamangir}\thanks{Corresponding Author} \\
	Department of Biological and Agricultural Engineering\\
	University of California Davis\\
        Davis, CA, USA\\
	\texttt{hkamangir@ucdavis.edu} \\
	\And
	{\includegraphics[scale=0.06]{orcid.pdf}\hspace{1mm}Brent. S. Sams} \\
	Department of Winegrowing Research\\
	E\&J Gallo Winery\\
	Modesto, CA, USA\\
	\texttt{Brent.Sams@ejgallo.com} \\
        \And
	{\includegraphics[scale=0.06]{orcid.pdf}\hspace{1mm}Nick Dokoozlian} \\
	Department of Winegrowing Research\\
	E\&J Gallo Winery\\
	Modesto, CA, USA\\
	\texttt{Nick.Dokoozlian@ejgallo.com} \\
        \And
	{\includegraphics[scale=0.06]{orcid.pdf}\hspace{1mm}Luis Sanchez} \\
	Department of Winegrowing Research\\
	E\&J Gallo Winery\\
	Modesto, CA, USA\\
	\texttt{Luis.Sanchez@ejgallo.com} \\
        \And
	{\includegraphics[scale=0.06]{orcid.pdf}\hspace{1mm}J. Mason. Earles} \\
	Department of Viticulture and Enology and Department of Biological and Agricultural Engineering\\
	University of California Davis\\
	Davis, CA, USA\\
	\texttt{jmearles@ucdavis.edu} \\ 
}

\renewcommand{\shorttitle}{CMAViT Climate Management Aware Vision Transformers Crop Yield Estimation}

\begin{document}
\maketitle

\begin{abstract}
Crop yield prediction is essential for agricultural planning but remains challenging due to the complex interactions between weather, climate, and management practices. To address these challenges, we introduce a deep learning-based multi-model called Climate-Management Aware Vision Transformer (CMAViT), designed for pixel-level vineyard yield predictions. CMAViT integrates both spatial and temporal data by leveraging remote sensing imagery and short-term meteorological data, capturing the effects of growing season variations. Additionally, it incorporates management practices, which are represented in text form, using a cross-attention encoder to model their interaction with time-series data. This innovative multi-modal transformer tested on a large dataset from 2016-2019 covering 2,200 hectares and eight grape cultivars including more than 5 million vines, outperforms traditional models like UNet-ConvLSTM, excelling in spatial variability capture and yield prediction, particularly for extreme values in vineyards. CMAViT achieved an R² of 0.84 and a MAPE of 8.22\% on an unseen test dataset. Masking specific modalities lowered performance: excluding management practices, climate data, and both reduced R² to 0.73, 0.70, and 0.72, respectively, and raised MAPE to 11.92\%, 12.66\%, and 12.39\%, highlighting each modality's importance for accurate yield prediction. Code is available at \url{https://github.com/plant-ai-biophysics-lab/CMAViT}. 
\end{abstract}


\section{Introduction}
Crop yield is a fundamental determinant of agricultural profitability, with early-season yield forecasting playing a critical role in guiding agronomic planning decisions, including procurement, sales, labor allocation, and essential management practices such as fertilization, irrigation, pest control, and soil and canopy management. Accurate yield forecasts enable effective planning and resource allocation, while inaccurate estimates can result in significant economic losses, environmental strain, and disruptions to the agricultural supply chain. 

\begin{figure}[ht]
\centering
\captionsetup{font=footnotesize}%
\includegraphics[width=9cm, height=5.5cm]{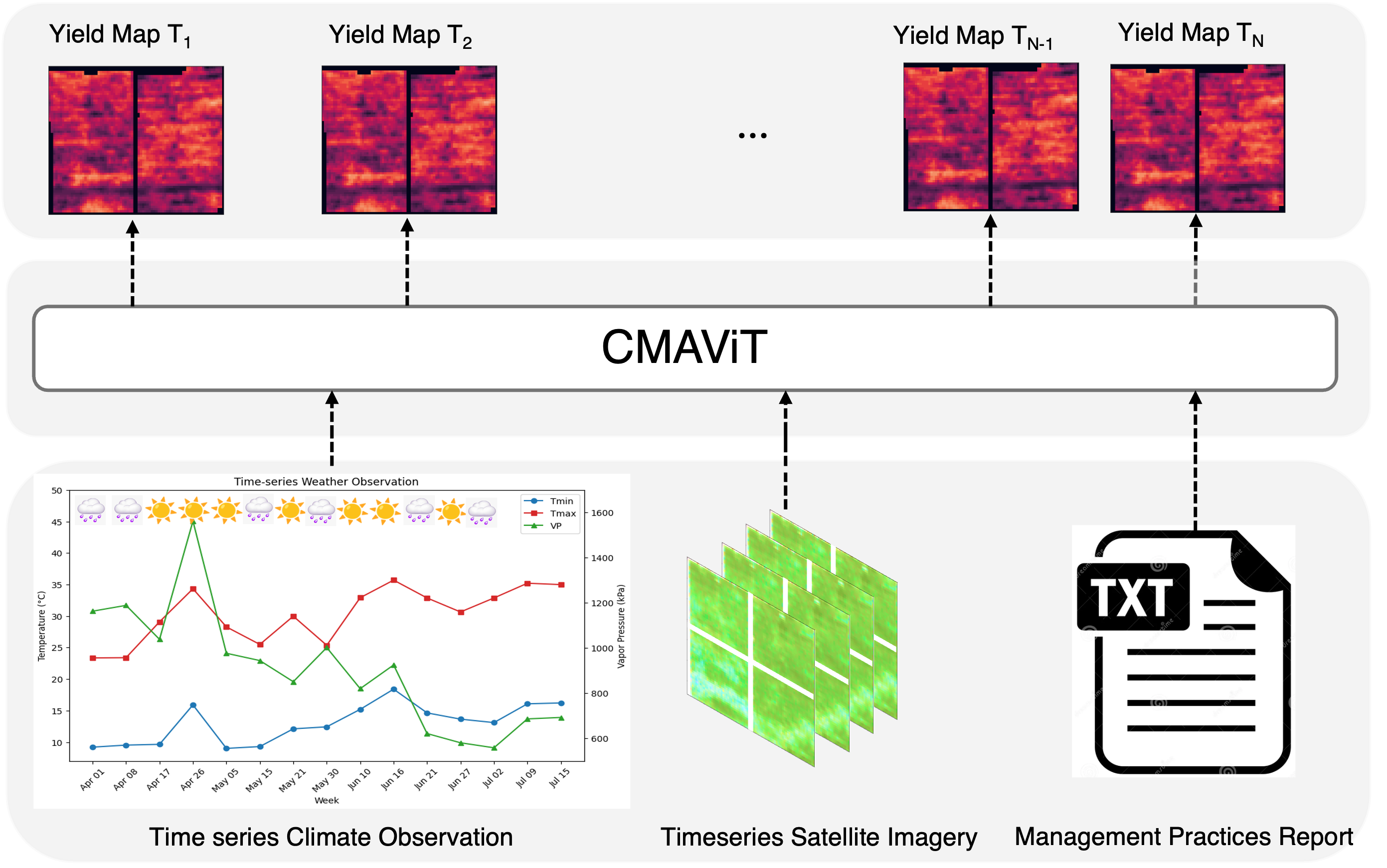} 
\caption{CMAViT: Climate, Management Aware Vision Transformer Model. A multi-model analysis integrating remote sensing imagery, climate data, and management practices to produce a time-series yield map throughout the flowering period, up to pre-harvest.}
\label{fig:init}
\end{figure}

Yield is influenced by a complex interplay of factors characterized by multivariate, non-linear, temporal, and spatial relationships. These factors include site-specific conditions such as climate, soil properties, topography, microclimate, pest pressures, and management practices, all of which contribute uniquely to yield outcomes \citep{jiang2004topoyield, CEGLAr201658, Epule2018, Ortiz_Bobea_2019, Chaloner2021}. Additionally, plant traits like canopy vigor, stress responses, flowering, and fruit development, observed early in the growing season, serve as indicators of potential yield by reflecting the cumulative effects of these factors \citep{Johnson2014CornSoyYield, liu2017computer, rudolph2019efficient, bai2019improving}. The inherent non-linearity of these interactions, combined with spatial variability, makes accurate yield prediction highly challenging, particularly given the annual fluctuations driven by genetic, environmental, and management interactions.

Historically, yield estimation for specialty crops has relied on collecting historical data and adjusting it based on current growing season conditions, such as weather, pest pressure, and specific management actions. Growers commonly sample a small number of plants to measure characteristics like cluster weight and number of clusters, using these observations to extrapolate the overall yield \citep{dami2006methods}. While widely used, this approach is labor-intensive, costly, and prone to inaccuracies due to the reliance on limited sample sizes that may not adequately represent field conditions. In contrast, modern research has increasingly focused on employing remote sensing and ground-based imagery to enhance yield forecasting accuracy through automation and high-resolution monitoring \citep{malik2016detection, gandhi2016rice, sun2017daily, di2019precision, liu2019grapeSVM, santos2020grape, liu2020vision, wei2020soybean, panek2021relationship, falco2021influence, cheng2022high, kamangir2024large}. Satellite and UAV (unmanned aerial vehicle) imagery have significantly advanced yield prediction by enabling frequent, detailed data collection across large areas, providing critical insights into crop growth and health \citep{Johnson2014CornSoyYield, liu2017computer, rudolph2019efficient, bai2019improving}. However, while remote sensing provides valuable top-of-canopy information, it cannot fully capture the intricacies of yield determinants, such as soil properties, pest pressures, and management practices, which necessitates integrating complementary data sources for a more comprehensive understanding \citep{jiang2004topoyield, CEGLAr201658, Epule2018, Ortiz_Bobea_2019, Chaloner2021}. Combining remote sensing data with additional information—such as climate, weather patterns, and site-specific management—allows for a more holistic approach to yield forecasting, ultimately enhancing prediction accuracy and supporting sustainable agricultural practices.

Yield prediction models generally fall into two main categories: mechanistic and data-driven approaches, each with distinct advantages and challenges \citep{kamangir2024large}. Mechanistic models explicitly incorporate plant structural traits, environmental variables, and management practices, providing a means to model linear and non-linear relationships over space and time \citep{kamangir2024large}. These models can effectively represent complex interactions; however, they require extensive empirical calibration and often rely on predefined assumptions about variable relationships, which may not apply universally across different crops or growing conditions \citep{Keating2003APSIM}. This limitation is particularly evident for specialty crops, where calibration and validation data are often scarce. In contrast, data-driven models, particularly those leveraging deep learning, offer a more scalable approach by learning directly from large datasets without relying on predefined assumptions. Data-driven models, such as those based on deep neural networks (DNNs), random forests, and support vector machines (SVMs), can capture complex spatial, temporal, and non-linear relationships within agricultural datasets \citep{khaki2020cnn, khaki2021simultaneous, kamangir2024large}. Hybrid approaches that integrate data-driven models with mechanistic models offer an even more powerful solution, combining the interpretability of mechanistic models with the adaptability of deep learning techniques, thereby improving the overall accuracy of yield prediction \citep{deines2021million}.

The application of machine learning, particularly deep learning, in agricultural yield prediction has shown substantial promise for capturing complex temporal, spatial, and spectral dependencies in the data. Conventional models, such as linear regression, have been used extensively to relate vegetation indices like NDVI to yield outcomes but have struggled to fully account for the non-linear and spatio-temporal relationships present in agricultural systems \citep{sun2017daily}. In response, more sophisticated machine learning methods—including random forests, SVMs, and deep learning approaches such as convolutional neural networks (CNNs), graph neural networks (GNNs), and Transformers have been developed and deployed to improve yield forecasting \citep{lecun2015deep, aneja2018convolutional, dosovitskiy2020image, khaki2021simultaneous, tseng2021cropharvest, fan2022gnn, cheng2022high, lin2023mmst, kamangir2024large}. While these models have demonstrated significant improvements, conventional CNN architectures still face challenges in integrating heterogeneous data sources—such as remote sensing imagery, climate data, and management practices—due to their rigid structures. These limitations have driven the development of hybrid models that incorporate subsets of available data sources, but they often lack a holistic representation of the diverse yield-influencing factors \citep{maaloy2021multimodal, zhou2021crop, kamangir2024large}. This gap underscores the need for a comprehensive multimodal approach capable of integrating multiple structured and unstructured data types to further improve yield prediction accuracy.

In this study, we propose a novel multimodal deep learning model for vineyard crop yield prediction at the field level, integrating various data types to enhance prediction accuracy. Our model incorporates Sentinel-1 and Sentinel-2 satellite data, capturing detailed imagery to provide insights into vegetation health, growth dynamics, and phenological development. Additionally, we integrate climate data from the DayMet dataset, encompassing variables such as minimum and maximum temperatures, precipitation, and vapor pressure, to accurately account for the effects of climatic variability on crop growth. We also include management and soil information from text-based reports, such as soil sampling results and descriptions of key management practices, allowing for a more comprehensive understanding of site-specific conditions. By combining these diverse data sources, our model aims to capture the complex interactions among environmental factors, management practices, and plant responses, thereby providing field-specific, accurate yield forecasts. This holistic approach is designed to support informed decision-making in vineyard management, enhance productivity, and promote sustainable agricultural practices.

\section{RELATED WORKS}\label{ch3:sec:background}

\subsection{Single-Modality Agricultural Crop Yield Prediction}

Many machine learning and deep learning models have been developed for crop yield prediction, typically based on a single modality, such as satellite imagery, ground-based imagery, or meteorological observations. The most common machine learning models used for crop yield forecasting include Random Forest \citep{everingham2016accurate, prasad2021crop, joshi2023winter, kuradusenge2023crop, bharadiya2023forecasting, amankulova2023comparison, panigrahi2023machine}, SVM \citep{liu2019grapeSVM, fegade2020crop}. However, these models often suffer from several limitations, including the need for extensive feature engineering, insufficient ability to capture non-linear relationships, and a lack of spatial correlation representation across fields and crop areas compared to more sophisticated deep learning architectures.

To overcome these limitations, deep learning models with layers of non-linear activation functions and specialized modules have been developed to better learn the spatial and temporal correlations and the complex interactions between predictors and yield outcomes. Models like Convolutional Neural Networks (CNNs) \citep{khaki2021simultaneous}, Graph Neural Networks (GNNs) \citep{fan2022gnn}, Long Short-Term Memory networks (LSTMs) \citep{kamangir2024large}, and Vision Transformers (ViTs) \citep{dosovitskiy2021imageworth16x16words} have emerged as effective tools for capturing both spatial and temporal relationships in agricultural data. These models have been successfully applied to predict crop yield at various scales, including field-level, county-level, and global-level. Their ability to learn non-linear features and model interactions between input variables has significantly improved yield prediction accuracy compared to traditional machine learning models.

In recent years, researchers have also explored the use of hybrid models that combine different deep learning architectures to further enhance yield prediction. These hybrid approaches leverage the strengths of multiple model types to better capture the diverse aspects of agricultural data. For example, combining CNNs with LSTMs allows for both spatial feature extraction and temporal sequence learning, making these models particularly well-suited for handling spatio-temporal data like satellite imagery over time \citep{kamangir2024large}. Similarly, the use of GNNs can help capture complex relationships between spatially connected regions, while Transformers provide flexibility in handling various input types and improving feature integration across different data sources \citep{khaki2021simultaneous}.

These advancements underscore the importance of moving beyond single-modality models to multimodal and hybrid approaches, enabling more accurate and reliable crop yield forecasting that can support informed decision-making in agriculture. Kamangir et al., 2024 \citep{kamangir2024large} developed the UNet-ConvLSTM model specifically for predicting spatio-temporal vineyard yields using solely Sentinel-2 imagery. This innovative model incorporates variables related to management practices, such as cultivar type, trellis type, spacing between rows, and canopy management, allowing it to learn and adapt to specific management strategies within each cultivar and field block. Despite its advancements, this model does not integrate weather data, which is a significant limitation given the impact of climatic conditions on crop yields. Additionally, the management practices are tailored uniquely to each cultivar type, and the model does not incorporate further details about variations within individual blocks, which could provide deeper insights into localized yield determinants.

\subsection{Multimodel Agricultural Crop Yield Prediction}

The world we live in is inherently multimodal, influencing both our observations and behaviors. Multimodal learning (MML) has been a significant research area in recent decades, with early applications like audio-visual speech recognition studied in the 1980s \citep{yuhas1989integration}. Furthermore, in the era of deep learning neural networks, transformers \citep{vaswani2017attention}, as a powerful and versatile architecture, present new challenges and opportunities to MML. The recent achievements of large language models and their multimodal variants highlight the potential of transformers, including vision transformers, in learning from diverse data sources.

Recent research has focused extensively on multimodal learning approaches for agricultural crop yield forecasting, leveraging diverse data sources to improve prediction accuracy \citep{maimaitijiang2020soybean, maaloy2021multimodal, kaur2022generalized, mia2023multimodal, ramzan2023multimodal, lin2023mmst}. These studies explore various combinations of data modalities, such as satellite imagery, weather information, soil data, and even genetic data, utilizing a variety of machine learning and deep learning techniques. Most of the recent efforts have centered on combining multiple sources of remote sensing data with different spatial resolutions \citep{ramzan2023multimodal}, integrating satellite and weather data \citep{mia2023multimodal, lin2023mmst}, and combining satellite, weather, and in-depth soil information using multimodal models like CNN-based architectures and Vision Transformers \citep{kaur2022generalized}. These advancements in multimodal fusion aim to enhance the predictive capabilities of crop yield models by capturing the complex interplay of environmental, biological, and management factors.

The study by \citep{maimaitijiang2020soybean} demonstrates the effectiveness of integrating UAV-based multispectral, thermal, and RGB imagery for predicting soybean yield using deep learning models. The research highlights that by combining spectral, structural, thermal, and texture features, prediction accuracy can be significantly improved. The study explored several regression techniques, including Partial Least Squares Regression (PLSR), Random Forest Regression (RFR), Support Vector Regression (SVR), and two deep neural network-based fusion models (DNN-F1 and DNN-F2). The deep learning model DNN-F2 outperformed others, achieving an R² of 0.720 and a relative RMSE of 15.9\%, indicating that multimodal fusion reduces the sensitivity to saturation effects commonly faced by remote sensing data in dense crop canopies. This work demonstrates that multimodal data fusion using UAVs is a viable approach for high-resolution, field-specific yield prediction, supporting precision agriculture and high-throughput plant phenotyping.

Maaloy et al, 2021 \citep{maaloy2021multimodal} introduces an innovative approach using transformer-based models known as Performers for predicting barley yield. This model integrates single nucleotide polymorphism (SNP) genotype data with weather data to capture the intricate relationships between genotype and environmental factors. Traditional genomic selection (GS) methods, such as Bayesian models or classical neural networks, often face challenges when modeling these complex interactions. However, the Performer model utilizes a self-attention mechanism, which requires fewer assumptions and is well-suited to learning from multimodal inputs. Despite these advances, the model's capabilities are limited to two types of data, restricting its ability to integrate other essential data sources like soil and management practices.

In a further advancement, \citep{kaur2022generalized} developed CropYieldNet, a multimodal deep learning model designed to incorporate spatiotemporal meteorological data, surface reflectance bands from satellite imagery, and depth-dependent soil information for predicting crop yield. CropYieldNet utilizes Convolutional Neural Networks (CNNs) to extract spatial features from reflectance data, combined with Long Short-Term Memory (LSTM) networks for temporal sequence learning. The introduction of a depth-level selection module significantly improved the model's performance, allowing it to effectively account for soil attributes, which are crucial for understanding crop growth dynamics. The model achieved up to a 7.8\% improvement in predictive accuracy compared to existing models, highlighting the value of integrating multiple modalities to capture the complexities inherent in crop yield prediction.

The research by \citep{mia2023multimodal} takes a similar approach, focusing on integrating UAV-based multispectral imagery with weather data to predict rice yield, particularly in regions dominated by smallholder farming. This multimodal deep learning model effectively combined spatial data from UAV imagery at the heading stage with temporal weather data to enhance prediction accuracy. The study evaluated different CNN architectures, including AlexNet and a simpler two-layer CNN, and found that while prediction accuracy was similar for both architectures, the simpler model was computationally more efficient. A key finding was that adding an additional layer after concatenating spatial and temporal features improved prediction accuracy, although the robustness of the model still requires further evaluation.

\begin{figure*}[ht]
\centering
\captionsetup{font=footnotesize}%
\includegraphics[width=16cm, height=10cm]{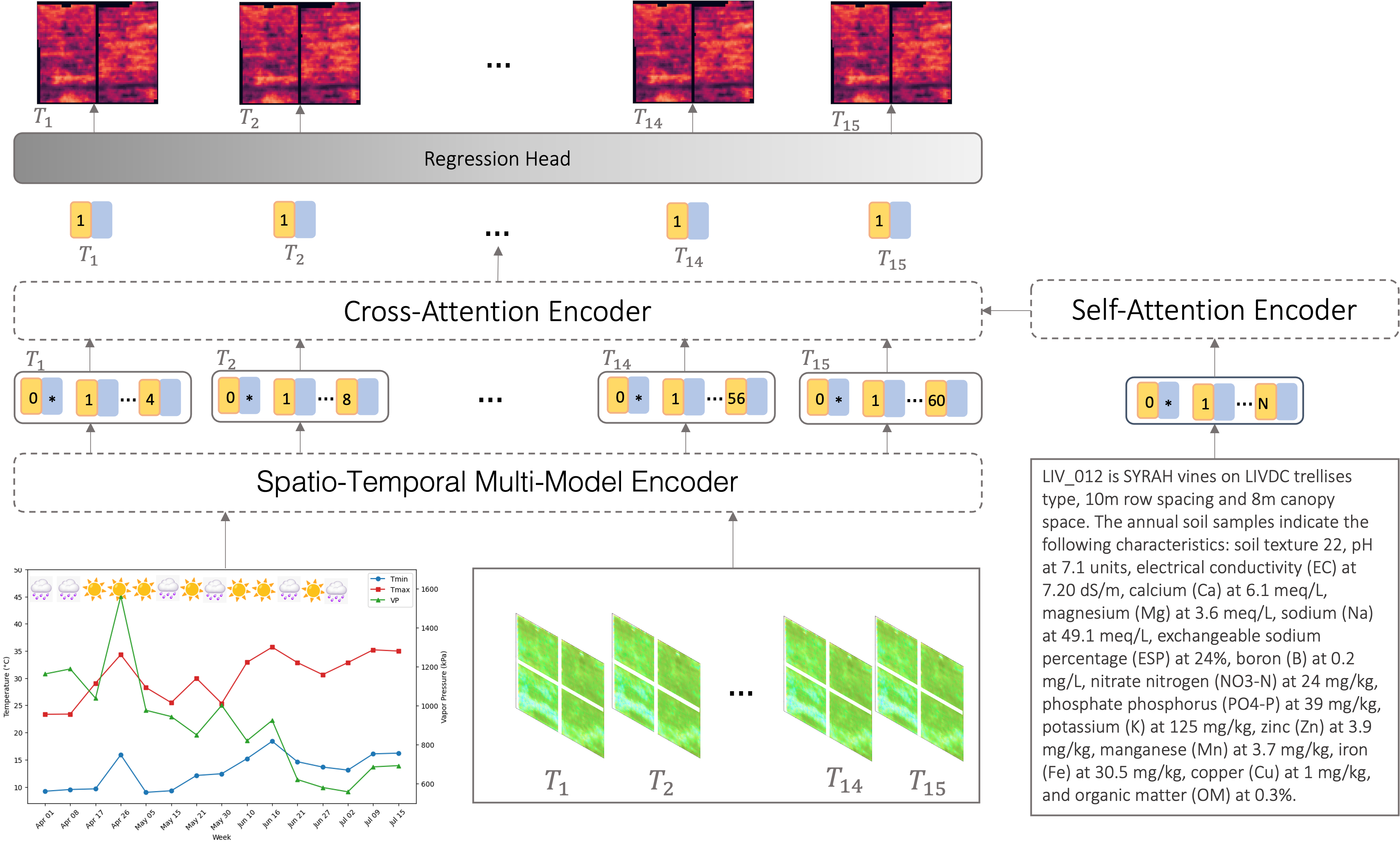} 
\caption{CMAViT: Climate, Management Aware Vision Transformer Model. The model employs a spatio-temporal multi-model framework that learns from time-series observations, including Sentinel-1 and Sentinel-2 satellite imagery, along with climate data. A cross-attention module captures interactions between the output from learned time-series features of the spatio-temporal module and contextual data from management practices. This enables weekly yield map predictions from bud break in early April through to veraison in mid-July.}
\label{fig:model}
\end{figure*}

In another example of multimodal learning, \citep{ramzan2023multimodal} investigated tea yield estimation in Pakistan by combining agrometeorological data with remote sensing imagery from Landsat-8. The proposed methodology used features such as NDVI from multispectral imagery, along with agrometeorological parameters, to train various machine learning and deep learning models. The best-performing model, a deep neural network (DNN) constructed using Bayesian optimization, achieved an impressive R² of 0.99 and demonstrated lower prediction errors compared to traditional machine learning algorithms like Decision Trees, Support Vector Regression (SVR), and Random Forest. This study underscores the potential of integrating agrometeorological and remote sensing data to accurately capture non-linear relationships and complex interactions in crop yield prediction.
Most recently, \citep{lin2023mmst} proposed a multimodal approach that combines remote sensing imagery with meteorological data for crop yield estimation using a Vision Transformer model pre-trained on long-term weather data. This climate-aware vision transformer achieved higher performance compared to benchmark models, effectively leveraging both spatial and temporal features. 

Overall, these studies collectively demonstrate the progress in multimodal learning for crop yield prediction, with significant advancements in integrating diverse data sources to capture the complexity of agricultural environments. Earlier models often focused on single modalities—whether satellite imagery, UAV-based data, or weather information—limiting their ability to fully understand yield determinants. By combining different types of spatial and temporal data, as well as incorporating additional data such as soil properties, management practices, and genetic information, recent models have shown significant improvements in accuracy and generalizability. Deep learning techniques like CNNs, LSTMs, and Vision Transformers have played a pivotal role in this evolution, enabling models to effectively learn the interactions among multiple predictors.

However, there remain several challenges in yield prediction research. The need for more comprehensive multimodal models that integrate structured and unstructured data—such as weather patterns, soil data, management practices, and satellite imagery—is becoming increasingly apparent. To address this gap, the authors propose developing a spatio-temporal multimodal vision transformer capable of incorporating various data types, such as images, weather data, and text-based management reports. Such a model could process input data temporally, offering more comprehensive predictions throughout the growing season.

\section{PROPOSED METHOD} \label{ch3:sec:Methodology}

This work aims to develop a model based on vision transformers for predicting vineyard yields at the field level through time-series analysis. The architecture shown in Figure~\ref{fig:model} is threefold: First, the Spatio-Temporal Multimodel (STMM) Vision Transformers capture the spatio-temporal signals from satellite imagery and integrate these with time-series climate conditions for enhanced lead-time predictions. Second, a self-attention encoder processes text-based input data, such as soil health and management practices, using a self-attention mechanism to encode pertinent information. Third, a temporal cross-attention mechanism employs cross-attention to assess the influence of the encoded text data on the spatial-temporal features for each prediction. This integrated approach leverages the strengths of vision transformers to provide precise, contextually aware yield forecasts.

\begin{figure}[ht]
\centering
\captionsetup{font=footnotesize}%
\includegraphics[width=5cm, height=6cm]{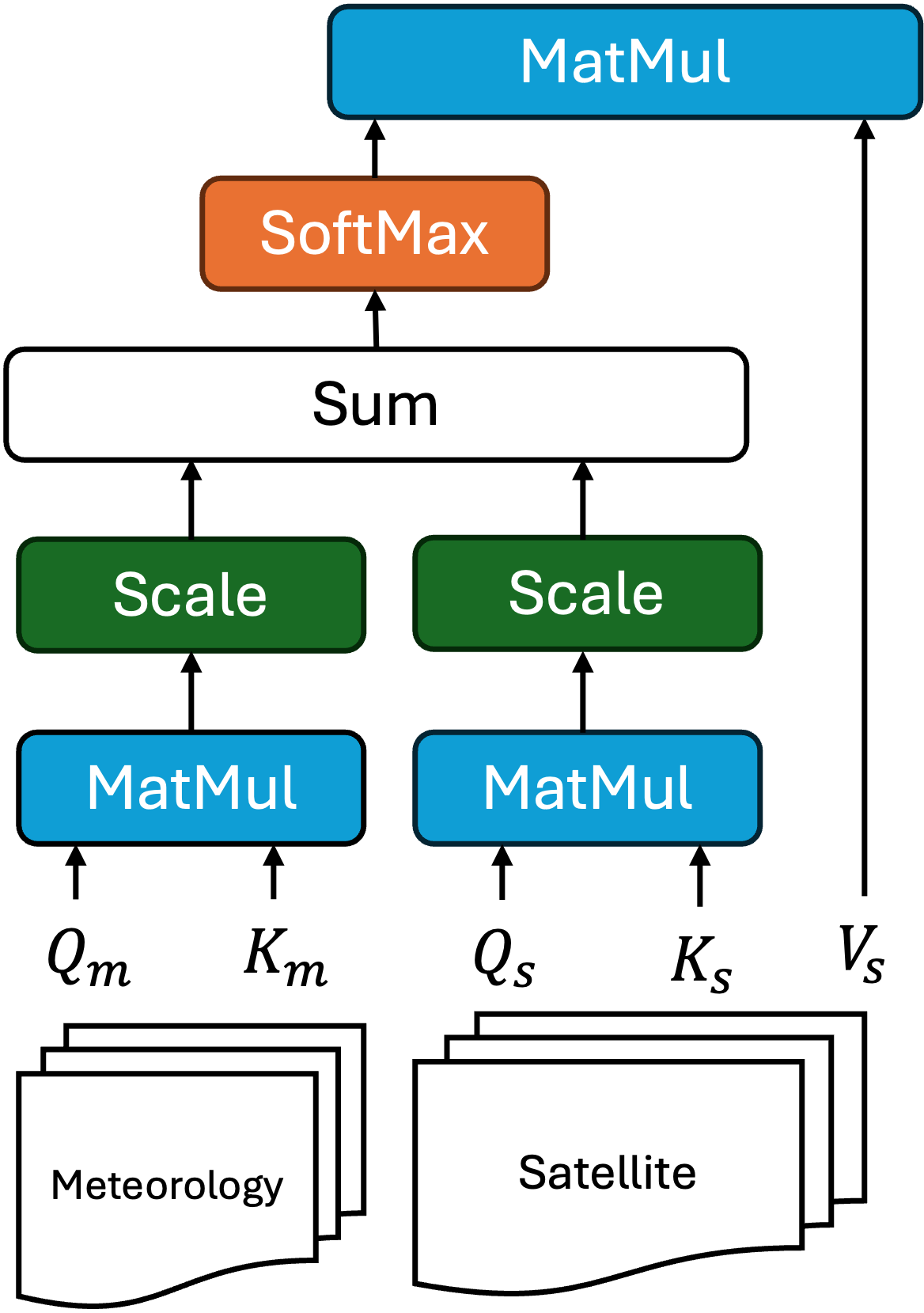} 
\caption{STMM Module. Multimodel to integrate meteorological observation with satellite imagery.}
\label{fig:stmm}
\end{figure}

\textbf{Capturing Crop Growth Responses to Weather Variation During the Growing Season} 
The STMM module (Figure~\ref{fig:stmm}) is designed to capture the impact of weather variation on crop growth by utilizing both visual remote sensing and numerical meteorological data. This module employs a sophisticated multi-head attention technique that systematically combines these two types of data. Specifically, the model integrates meteorological signals with satellite imagery in an accumulative manner, enhancing its ability to predict crop responses for each lead time.

By doing so, it ensures that the temporal dynamics of weather conditions are effectively reflected in the spatial analysis provided by satellite observations. This integrated approach would improves the accuracy of predictions and provides deeper insights into how different weather patterns influence crop growth over time. 

\begin{equation}\begin{aligned}
    STMM(Q_s, K_s, V_s, Q_m, K_m) = &\quad \\ 
    \text{softmax} \left( \frac{Q_s K_s^T}{\sqrt{d}} + \frac{Q_m K_m^T}{\sqrt{d}} \right) V_s
 \end{aligned}\end{equation}

\begin{equation}
    \begin{aligned}
        Q_s &= W_s^Q \cdot \phi_s(x), &\quad  
        K_s &= W_s^K \cdot \phi_s(x), &\quad \\ 
        V_s &= W_s^V \cdot \phi_s(x), &\quad 
        Q_m &= W_m^Q \cdot \pi_m(met), &\quad \\ 
        K_m &= W_m^K \cdot \pi_m(met), &\quad
    \end{aligned}
\end{equation}

Here, $\phi_s(x) \in \mathbb{R}^{(T \times N_p) \times d}$ is the visual representation encoded by a linear projection, with $T$ indicating the temporal observation and $N_p$ the number of image patches. $\pi_m(met) \in \mathbb{R}^{(T \times N_v) \times d}$ represents the meteorological parameters after the linear projection, with $N_v$ indicating the number of meteorological values. Notably, $W_s^Q$, $W_s^K$, and $W_s^V$ are learnable projection matrices. Specifically, for satellite imagery, we denote the data as \( x \in \mathbb{R}^{T \times H \times W \times C} \), where \( T \) represents the number of temporal observations, and \( C \) represents the number of channels. In this experiment, the satellite images are composites of Sentinel-1 and Sentinel-2, incorporating RGB and NIR channels from Sentinel-2 and VV and HV channels from Sentinel-1, totaling six channels. The embedding patches for satellite imagery is:

\begin{equation}
    \phi_s(x)^t = 
    \begin{aligned}
        &\left[ v_s^{cls}; v_s^{N_1, t_1}; v_s^{N_2, t_1}; v_s^{N_3, t_1}; v_s^{N_4, t_1}; \right. \\
        &\left. v_s^{N_1, t_2}; v_s^{N_2, t_2}; \ldots; v_s^{N_4, T} \right] + \mathbf{E}_{\text{tmp}},
    \end{aligned}
\end{equation}

Where $\mathbf{E}_{\text{tmp}} \in \mathbb{R}^{(T*N_p+1) \times d}$ is the temporal embedding. 
For the numerical weather data, we denote the data as \( met \in \mathbb{R}^{T \times 1 \times C} \) the temporal observations align with those of the satellite imagery, and \( C \) is four, representing Tmin, Tmax, Prcp, and VP.

\textbf{Capturing Soil Health and Management Practices Information} 

This study employs a vanilla self-attention encoder to effectively learn from text-based information regarding soil health and management practices using pre-trained GPT-3.5 tokenizer \citep{openai2023gpt3.5}. By harnessing this encoder, we aim to transform complex textual data—ranging from soil composition reports to detailed management logs—into a structured format that can be seamlessly integrated into larger agricultural models. This approach not only simplifies the processing of extensive textual data but also ensures that critical information influencing agricultural outcomes is accurately captured and utilized, thus enhancing the overall understanding and strategic planning of soil health management.

\textbf{Capturing Soil Health and Management Practices Impacts on Crop Growth During the Growing Season} 
In this research, we utilize a cross-attention module to analyze time-series outputs for predicting crop growth impacts during the growing season, considering variations in soil health and management practices. This module allows for a dynamic assessment of how different soil and management variables interact over time, focusing particularly on their temporal influence on crop development. By integrating cross-attention mechanisms, the model adeptly correlates fluctuating soil conditions and management actions with plant growth patterns, providing nuanced insights into the temporal dynamics that drive agricultural productivity throughout the season.

\begin{equation}
    CA(Q_{x}, K_t, V_t) = \text{softmax} \left( \frac{Q_{x} K_t^T}{\sqrt{d}} \right) V_t
\end{equation}

\begin{equation}
    \begin{aligned}
        Q_{x} &= W_{x}^Q \cdot \phi_s(x), & \quad 
        K_t &= W_t^K \cdot \phi_s(context), & \quad \\
        V_t &= W_t^V \cdot \phi_s(context), 
    \end{aligned}
\end{equation}

Here, $\phi_s(x) \in \mathbb{R}^{(T \times N_p) \times d}$ is the visual representation encoded by a linear projection, with $T$ indicating the temporal observation and $N_p$ the number of image patches. $\pi_m(met) \in \mathbb{R}^{(T \times N_v) \times d}$ represents the meteorological parameters after the linear projection, with $N_v$ indicating the number of meteorological values. Notably, $W_s^Q$, $W_s^K$, and $W_s^V$ are learnable projection matrices.

\section{EXPERIMENTS AND DISCUSSION}

\subsection{Datasets} 
\subsubsection{Vineyard Yield Observation}
\begin{figure*}[ht]
\centering
\captionsetup{font=footnotesize}%
\includegraphics[width=16cm, height=8cm]{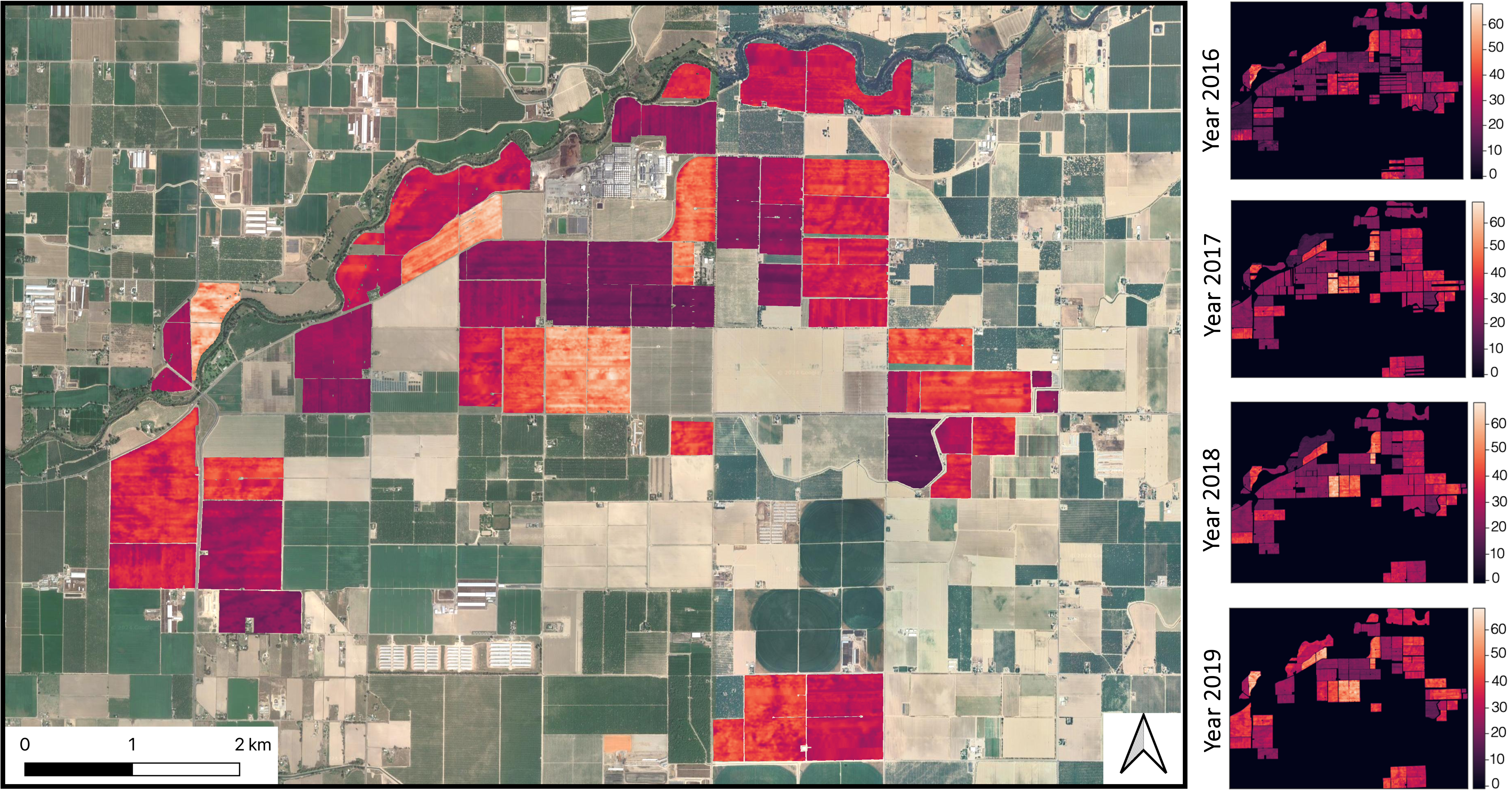} 
\caption{Four year yield observations tonne/hectare (t/ha) ground-truth dataset in the Central Valley of California collected by E and J Gallo Winery. Each map is about 7 x 9 km containing about 2,200 hectare of over 5 million grapevine plants.}
\label{fig:area}
\end{figure*}

In this study, we leverage E and J Gallo Winery's ongoing efforts to collect high-resolution (1-3 m) yield data via monitors on harvesters (Advanced Technology Viticulture, Joslin, South Australia, Australia) in California's Central Valley (Figure \ref{fig:area}). Spanning 4 years (2016-2019), the dataset encompasses 7x9 km, covering 2,200 hectares, over 5 million grapevine plants, and 15 cultivars. This extensive ground-truth data allows for robust validation of yield estimation models, utilizing deep learning to decipher complex interactions between inputs and yield.

\subsubsection{Satellite Images}
We correlated the yield observation dataset with time-series imagery from the Sentinel-2A, Sentinel-2B, and Sentinel-1 satellites \citep{Sentinel2}, accessed through Google Earth Engine (GEE) \citep{gorelick2017google}. The selected images were captured on dates that were closely aligned in terms of the day of the year. Following the launch of Sentinel-2B, the temporal resolution of Sentinel-2 imagery improved to a 5-day revisit cycle. We chose weekly images (ranging from 5 to 8 days apart) with less than 10\% cloud cover, spanning from April to mid-July for the studied years. Early Sentinel-2 images were in TOA (Top-of-Atmosphere) reflectance, whereas later images were in BOA (Bottom-of-Atmosphere) reflectance, requiring atmospheric correction using Sen2Cor, based on the LIBRADTRAN model \citep{mayer2005libradtran}. For Sentinel-1 imagery, VV and VH channel images were denoised using non-local means filters \citep{buades2005non}, which identify similar patches within the dataset to optimally suppress speckle noise without sacrificing image resolution.

\subsubsection{Management Practices Data}

Management practices throughout the growing season, including details such as cultivar type, trellis configuration, canopy and row spacing, soil texture, and various chemical properties, have been documented in text format. These chemical properties encompass pH, sodium, magnesium, and other key soil nutrients and elements essential for vine growth. Such comprehensive data provides valuable context regarding vineyard management, facilitating a deeper understanding of how different factors, such as soil composition and cultivation techniques, influence crop yield and health. This text-based dataset ensures that critical agronomic information is preserved for further analysis and interpretation. 

\subsubsection{Climate Data}

As demonstrated, climate data significantly impacts the understanding of interactions between environmental changes and vineyard crop growth. In this experiment, weekly climate data was collected on the exact same days as the satellite imagery, focusing on four critical observations: minimum temperature (Tmin), maximum temperature (Tmax), precipitation (prcp), and vapor pressure (VP, in kPa). These data were sourced from the DayMet dataset \citep{https://doi.org/10.3334/ornldaac/2129}, which provides a spatial resolution of 1 km. Since each vineyard field is smaller than 1 km by 1 km, a single value was assigned to represent each field for every observation period. This consistent and synchronized collection of climate variables enables a more precise analysis of how weather conditions affect vineyard growth dynamics in tandem with satellite-based observations.

Data preprocessing for the deep learning model involved several crucial steps to ensure data quality and suitability for model training. First, images with more than 10\% cloud cover were filtered out using the s2cloudless dataset. From an initial set of 15 cultivars, we excluded those that had fewer than two blocks or less than three years of data. After this refinement process, the final dataset included 41 blocks representing 8 cultivars across 4 years. The selected cultivars were Cabernet Sauvignon (CS), Chardonnay (Ch), Malvasia Bianca (MB), Merlot (Me), Muscat of Alexandria (MoA), Riesling (Ries), Symphony (Sym), and Syrah (Syr). This curation process ensured that the dataset was representative and adequately structured for effective deep learning model training, focusing on consistent spatial and temporal characteristics across the selected cultivars.

\subsection{Validation Scenario and Evaluation Metrics}
\begin{figure*}[ht]
\centering
\captionsetup{font=footnotesize}%
\includegraphics[width=16cm, height=6cm]{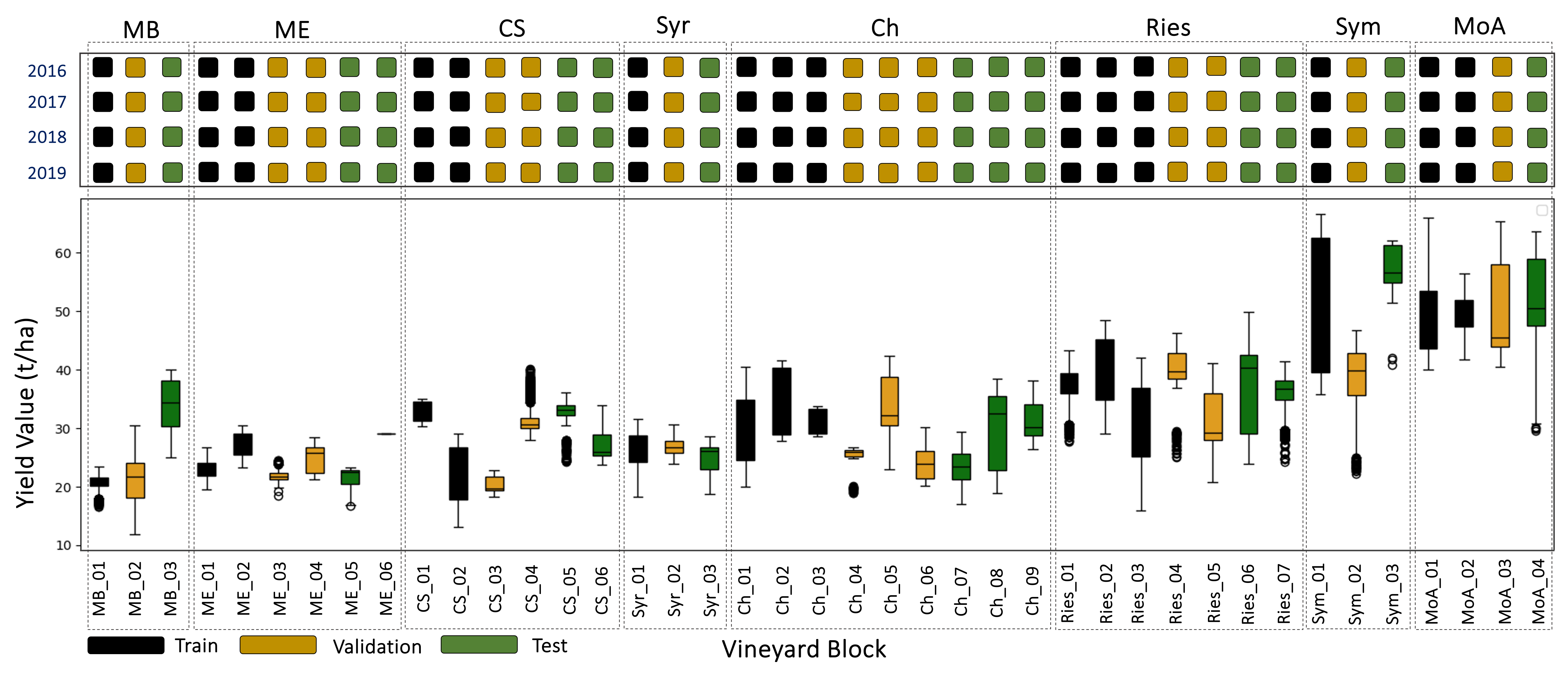} 
\caption{Block Holdout Validation Scenario (BHO). Malvasia Bianca (MB), Merlot (Me), Cabernet Sauvignon (CS), Syrah (Syr), Chardonnay (Ch), Riesling (Ries),  Symphony (Sym), and Muscat of Alexandria (MoA).}
\label{fig:bho}
\end{figure*}

To evaluate models generalization across new data blocks with varying cultivar types and management practices, we implemented validation scenario  called block-hold-out (BHO), depicted in Figure~\ref{fig:bho}. The scenario evaluates the model's ability to predict yields for blocks that were not part of the training set. 

This approach assesses how well the model generalizes to new blocks without historical yield data. For instance, in the MB cultivar type, one block was used for training, another for validation, and a third for testing. Similarly, for the Ch cultivar type, three out of nine blocks were assigned to each phase. The lower portion of the figure presents the boxplot distribution of historical yield data, showing yield variability within the same cultivar across different blocks, as well as between cultivars. This emphasizes the challenge of yield prediction due to the inherent variability in agriculture. 

To evaluate model performance, we employed four metrics, each highlighting different aspects of efficacy. The coefficient of determination ($R^2$) measures how well the model explains variance in the data, with higher values indicating better fit, although it can be susceptible to overfitting. Root Mean Square Error (RMSE) captures prediction accuracy by penalizing larger errors, making it useful for detecting major deviations but sensitive to outliers. Mean Absolute Error (MAE) offers a straightforward average of errors, treating all equally and being less influenced by outliers. Mean Absolute Percentage Error (MAPE) provides a relative measure of accuracy in percentage terms, which is helpful for comparing performance across different scales but can be unreliable when values are near zero. These metrics collectively ensure a thorough assessment, covering model fit, error magnitude, and consistency across varying conditions.

\subsection{Implementation Details} 
In this experiment, the images were cropped into patches of 16 by 16 pixels to account for spatial autocorrelation within the vineyard field. This resulted in data with dimensions of (T, C, H, W), where T represents time (15 time points), and C represents the number of channels—4 for Sentinel-2, 2 for Sentinel-1 imagery and 1 for time as an encoded day of the year of captured images using the sine transformation, given by 
\[
\text{DOY} = \sin\left(\frac{2 \pi \times \text{day\_of\_year}}{365}\right)
\]
to capture the cyclical nature of the growing season. Climate data was structured as (T, C, 1), with T also equal to 15 and C representing four climate variables. 

The experiments were all implemented with PyTorch on a single NVIDIA TITAN RTX GPU with 24 GB RAM. The models are trained over 300 epochs, with early stopping implemented at 50 epochs, using the AdamW optimizer \citep{loshchilov2017decoupled}. We set the optimizer’s parameters as $\beta_1 = 0.98$, $\beta_2 = 0.95$, with a weight decay of 0.01 and an initial learning rate of 1e-4. The architecture includes a self-attention module composed of 8 attention heads and 6 layers, featuring an embedding size of 768. The MLP (multi-layer perceptron) module has a hidden layer size of 2048, and a dropout rate of 30\% is applied to prevent overfitting. The context embedding also have a size of 768, and text tokenization is performed using pretrained GPT-3.5 \citep{openai2023gpt3.5}, which enhances the model’s ability to process textual input effectively. These configurations are designed to balance model capacity with regularization to achieve robust performance over time.

\subsection{Performance Comparison} \label{QResults}
This section presents the performance of the CMAViT model compared to the UNet-ConvLSTM model developed by \citep{kamangir2024large} across the entire dataset, time-series prediction, learning of extreme values, and spatio-temporal accuracy.

\subsubsection{Quantitative Results Across the Entire Dataset}

\begin{table*}[ht!]
\captionsetup{font=footnotesize}%
\footnotesize
\centering
\caption{Evaluation of CMAViT and UNet-ConvLSTM performance across various input configurations in the BHO validation scenario. Metrics include R-squared (R²), Mean Absolute Error (MAE $t/ha$), Root Mean Squared Error (RMSE), and Mean Absolute Percentage Error (MAPE\%). [A] Sentinel-2 imagery with management context; [B] Sentinel-2 imagery, and management context with CSR; [C] Sentinel-1 and Sentinel-2 imagery, with management context; [D] Sentinel-1, Sentinel-2, climate data, and management context; and [E] Sentinel-1, Sentinel-2, climate data, management context, and CSR. }
\begin{tabular}{l p{0.4cm} p{0.45cm} p{0.65cm} p{1.2cm}  p{0.4cm} p{0.45cm} p{0.65cm} p{1.2cm} p{0.4cm} p{0.45cm} p{0.65cm} p{1.1cm}} 
 \toprule
    & \multicolumn{4}{c}{Train} & \multicolumn{4}{c}{Validation} & \multicolumn{4}{c}{Test} \\
    \cmidrule(r){2-5} \cmidrule(r){6-9} \cmidrule(r){10-13}

Model & $R^{2}$ & MAE & RMSE & MAPE(\%) & $R^{2}$ & MAE & RMSE & MAPE(\%) &$R^{2}$ & MAE & RMSE & MAPE(\%)\\ 
 \hline\hline
UNet-ConvLSTM [A]  &  0.72 & 4.52 & 6.12 & 14.47 &  0.66 & 3.89 & 5.22 & 14.76 & 0.65 & 4.24 & 5.28 & 12.72\\ 
UNet-ConvLSTM [B] &  0.88 & 3.25 & 4.53 & 9.99 & 0.71 & 4.24 & 5.38 & 14.47 & 0.78 & 4.54 & 6.07 & 12.37\\ 
UNet-ConvLSTM [C] &  0.82 & 3.76 & 4.93 & 12.29 & 0.66 & 3.94 & 5.24 & 13.64 & 0.71 & 3.86 & 5.26 & 12.57\\ 
\hline
CMAViT [A]  &  0.77 & 4.19 & 5.51 & 13.34 & 0.59 & 4.33 & 5.71 & 14.77 &0.69 & 3.99 &5.52 & 12.61\\
CMAViT [B]  & 0.88 & 3.08 & 3.94 & 10.86 & 0.70 & 3.67 & 4.91 & 13.49 & 0.78 & 3.49 & 4.62 & 11.00 \\
CMAViT [C]  &  0.93 & 2.36 & 3.08 & 7.92 & 0.67 & 3.71 & 5.09 & 13.07 &0.74 & 3.73 & 4.98 & 11.48\\
CMAViT [D] &  0.94 & 2.14 & 2.84 & 7.34 & 0.67 & 3.82 & 5.15 & 13.87 &0.80& 3.29 & 4.43 & 10.19\\
CMAViT [E] & 0.93 & 2.24 & 2.96 & 7.26 &  0.83 & 2.74 & 3.72 & 9.26  & 0.84 & 2.77 & 3.90 & 8.22 \\\hline
UNet-ConvLSTM-YZ &  0.97 & 1.39 & 1.95 & 4.94 & 0.94 & 1.58 & 2.22 & 6.19 & 0.96 & 1.42 & 1.94 & 4.51\\
CMAViT-YZ &  0.97 & 1.34 & 1.89 & 4.70 & 0.94 & 1.51 & 2.12 & 6.02 & 0.95 & 1.52 & 2.25 & 4.45\\
\hline\hline
\label{table ch3 all}
\end{tabular}
\end{table*}

Table~\ref{table ch3 all} presents the performance of three categories of models across various input configurations for UNet-ConvLSTM and CMAViT. The first category is the UNet-ConvLSTM model developed by \citep{kamangir2024large}, the second pertains to the CMAViT model, and the third category compares these two models using the YieldZone strategy developed by \citep{kamangir4893816predicting}. The input configurations include the following categories: [A] Sentinel-2 imagery with management context; [B] Sentinel-2 imagery, and management context with CSR; [C] Sentinel-1 and Sentinel-2 imagery, with management context; [D] Sentinel-1, Sentinel-2, climate data, and management context; and [E] Sentinel-1, Sentinel-2, climate data, management context, and CSR.

The results presented in Table~\ref{table ch3 all} indicate a modest improvement in the performance of the CMAViT model compared to the CNN-based UNet-ConvLSTM model across all tested configurations. For example, the R² value improved from 0.65 to 0.69, and the MAPE  decreased from 12.72\% to 12.61\% for the UNet-ConvLSTM and CMAViT models, respectively, based on the data configuration A and unseen test dataset. Additionally, incorporating the CSR strategy (category B) led to a further reduction in MAPE, from 12.37\% to 11.00\% for test dataset.

The inclusion of Sentinel-1 data (category C) further enhanced the model performance for both architectures, with CMAViT showing superior results: an of 0.74 compared to 0.71 for UNet-ConvLSTM, and a MAPE of 11.48\% versus 12.57\% for unseen test dataset. The CMAViT model, when provided with the full input data including meteorological observations (category D), outperformed all other configurations, achieving an R² of 0.80 and a MAPE of 10.19\%. This highlights the significant impact of incorporating meteorological and climate data for improved crop yield prediction. The category E for full input observation alongside with CSR strategy improved the results for R² as 0.84 and MAPE 8.22\% for the unseen test dataset. 

The CMAViT model outperforms the UNet-ConvLSTM model in this context, and it also offers a more straightforward strategy for integrating diverse data structures, enhancing its ability to accurately predict yield values. Specifically, CMAViT effectively combines vector-based meteorological data, raster satellite imagery, and text-based management practice reports in this case study, demonstrating its versatility in handling complex datasets. The model's design is inherently open to incorporating larger and more intricate data inputs, making it highly adaptable for different agricultural applications.

On the other hand, while the UNet-ConvLSTM model also integrates satellite imagery with management information, it is limited by the number of categorical data points it can encode. In the UNet-ConvLSTM implementation, only four management features—cultivar type, trellis type, canopy space, and row space—are utilized, which restricts the richness of information that can be modeled. Conversely, CMAViT does not impose such limitations, as it allows the integration of management information in text format without predefined constraints, enabling the model to benefit from a broader and more detailed representation of farming practices. This flexibility in combining a variety of data types positions CMAViT as a more effective tool for yield prediction, especially in environments requiring comprehensive data integration.

The third category of results involves models based on the YieldZone strategy \citep{kamangir4893816predicting}, which demonstrates a significant improvement in learning extreme values and addressing the limitations of the MSE loss function, which tends to overemphasize learning from the majority of pixels. The YieldZone strategy initially segments the yield map into distinct categories based on three fundamental yield ranges—low-extreme, common, and high-extreme. However, the number of classes can vary depending on the specific task, data, and crop types. The output from this initial step is referred to as the "Yield Zone Map." The second step involves performing an element-wise multiplication of the Yield Zone Map with the input image data, which in this case is satellite images. However, this process is only deployed through training process not inference as it has been explained in Algorithm~\ref{alg:one}. 

\begin{algorithm}
\caption{Conditional training process of the crop yield prediction model}\label{alg:one}
\begin{algorithmic}[1]
\For{each batch}
    \State $X_0 \gets \text{get Sentinel-2 image}$
    \State $YZ \gets \text{classify crop yields into distinct classes}$
    \State $Z \gets \text{element-wise multiplication}(X_0, YZ)$
    \State Train model $M$ on $Z$
    \State Update $M$ to minimize loss $\mathcal{L}$
\EndFor
\end{algorithmic}
\end{algorithm}

Algorithm~\ref{alg:one} illustrates the entire process of training based on the YieldZone strategy. Phase 1, described as Algorithm~\ref{alg:one}, involves conditional training of the crop yield prediction model utilizing the YieldZone strategy. This phase processes batches of Sentinel-2 images (denoted as X) and classifies crop yields (denoted as Y) into a distinct number of classes (denoted as YZ). The model then performs element-wise multiplication of the X and 
YZ to produce Z, on which the model M is trained to minimize the loss 
$\mathcal{L}$. This conditional training process integrates spatial and class-specific information, optimizing the model specifically for varied yield zones. Phase 2 is the inference process, where the model input is solely the input image.

By applying the YieldZone strategy, the MAPE for the CMAViT model decreased significantly to 4.45\%, indicating a substantial enhancement in the model’s ability to accurately predict yield across different yield zones, particularly in cases involving extreme variations.

\subsubsection{Quantitative Results of Time Series Yield Estimation}
\begin{figure*}[ht]
\centering
\captionsetup{font=footnotesize}%
\includegraphics[width=16cm, height=8cm]{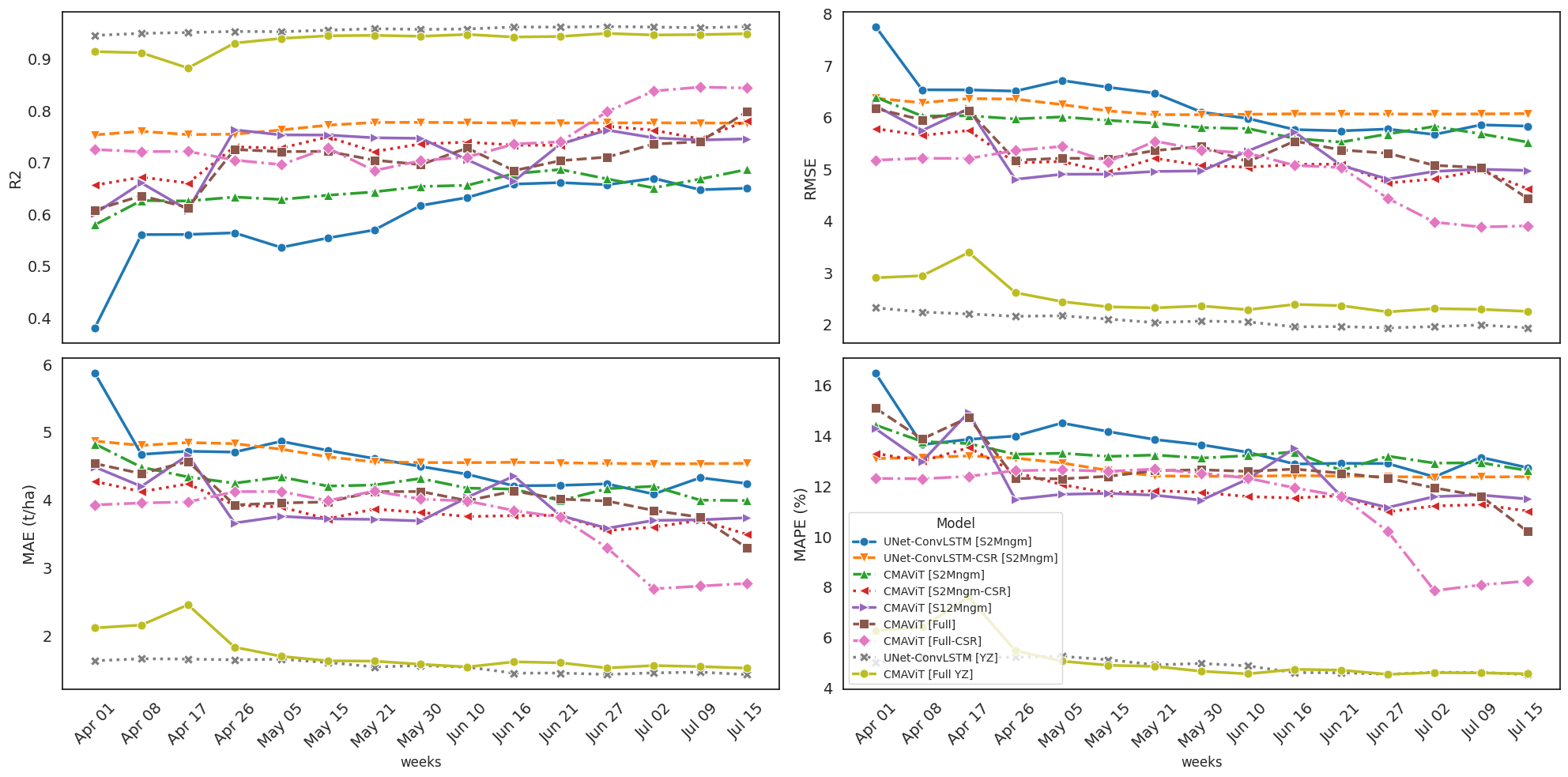} 
\caption{Time Series Forecasting Analysis Using UNet-ConvLSTM and CMAViT Models. [A] Sentinel-2 imagery with management context; [B] Sentinel-2 imagery, and management context with CSR; [C] Sentinel-1 and Sentinel-2 imagery, with management context; [D] Sentinel-1, Sentinel-2, climate data, and management context; and [E] Sentinel-1, Sentinel-2, climate data, management context, and CSR.}
\label{fig:ch3 timeseries}
\end{figure*}

The CMAViT model is designed to predict the time series of yield from early April, which corresponds to the flowering period, until mid-July, known as the veraison stage—a critical phase approximately one month before harvest. Based on the results presented in Figure~\ref{fig:ch3 timeseries}, all models exhibit a gradual improvement in prediction accuracy from week 1 through week 15, which aligns with the approach of the harvesting period. This trend suggests that as the growing season progresses, the models gain access to increasingly informative data, enabling them to better capture the complexities associated with yield prediction.

The improvement observed over the weeks underscores the importance of continuous data collection throughout the growing season. As plants progress through their phenological stages, the accumulation of data—such as changes in satellite imagery, updated meteorological inputs, and ongoing management practices—provides the model with richer and more representative information. This incremental data availability allows the model to refine its understanding of yield determinants, capturing dynamic growth factors and environmental influences more accurately. Consequently, this leads to enhanced predictive performance as the crop approaches maturity, where the variability in yield becomes more evident and can be modeled with greater precision.

The results presented in Figure~\ref{fig:ch3 timeseries} confirm the superior performance of CMAViT and CMAViT-CSR models throughout the prediction timeline, particularly in the final weeks leading up to the harvest. This indicates that the CMAViT models are more effective in capturing the underlying temporal dynamics as the harvest approaches. The changes observed over time are notably more pronounced for the CMAViT models compared to those based on the UNet-ConvLSTM architecture, suggesting a stronger capability of the CMAViT models in adapting to time-varying features and complexities inherent in agricultural processes.

Moreover, the results emphasize the improvements achieved by applying the yield zone strategy for both CMAViT and UNet-ConvLSTM models, which contributes to enhanced predictive accuracy. By incorporating the yield zone approach, the models effectively leverage spatial heterogeneity, leading to better representation of yield variations across different regions. This strategy appears to have a particularly positive impact on the CMAViT-based models, highlighting their suitability for addressing spatially complex tasks in agricultural forecasting.

\subsubsection{Quantitative Results of Extreme Yield Estimation}

Table~\ref{table:ch3 exresults} presents the results for the UNet-ConvLSTM and CMAViT-based models, both with and without the cost-sensitive resampling (CSR) and YieldZone (YZ) strategies over the entire range of yield values, only Low Extreme Range (LER, values less than 22 t/ha), Common Range (CR, between 22 and 54 t/ha) and High Extreme Range (HER, value higher than 54 t/ha). Overall, the CMAViT models demonstrate superior performance over the UNet-ConvLSTM models for predicting both low and high extreme values, based on MAE and MAPE evaluation metrics. Specifically, for models that utilize Sentinel-2 and management practices data (category A), the MAE across all samples improves from 4.50 to 3.99 t/ha, while MAPE decreases from 13.87\% to 12.61\%. In the low extreme range (LER), MAE is reduced from 5.68 t/ha to 5.01 t/ha, and MAPE improves from 29.98\% to 26.59\%. Similarly, for the high extreme range (HER), MAE significantly decreases from 17.53 t/ha to 9.02 t/ha, and MAPE is reduced from 28.98\% to 14.76\%.

\begin{table*}[ht!]
\captionsetup{font=footnotesize}%
\footnotesize
\centering
\caption{Evaluation of the yield zone strategy's impact on the performance of CMAViT and UNet-ConvLSTM models for extreme yield range values. LER = Low Extreme Range, CR = Common Range, HER = High Extreme Range. A: S2-Mngm, B: S2-Mngm-CSR, C: S12-Mngm, D: Full, E: Full-CSR.}
\begin{tabular}{l  p{0.9cm} p{0.9cm} p{0.9cm} p{0.9cm} p{0.05cm} p{0.9cm} p{0.9cm} p{0.9cm} p{0.9cm}} 

\toprule
\multicolumn{1}{c}{}& \multicolumn{4}{c}{MAE (t/ha)}& &\multicolumn{4}{c}{MAPE (\%)} \\\cmidrule{2-5} \cmidrule{7-10}
Model & All & LER & CR & HER &&All & LER & CR & HER\\
 \hline\hline
UNet-ConvLSTM [A] & 4.50 & 5.68 & 3.66 & 17.53 & & 13.87 & 29.98 & 11.56 & 28.98 \\ 
UNet-ConvLSTM [B] & 4.54 & 2.97 & 3.59 & 9.25 & & 12.37 & 15.46 & 11.49 & 15.13 \\ \hline
CMAViT [A]        & 3.99 & 5.01 & 3.61 & 9.02 & & 12.61 & 26.59 & 11.21 & 14.76 \\
CMAViT [B]        & 3.49 & 3.32 & 3.28 & 7.49 & & 11.00 & 17.76 & 10.32 & 12.27 \\
CMAViT [C]        & 3.73 & 3.19 & 3.46 & 9.45 & & 11.48 & 16.94 & 10.76 & 15.54 \\
CMAViT [D]        & 3.29 & 2.90 & 3.06 & 8.09 & & 10.19 & 15.16 & 9.52 & 13.35 \\
CMAViT [E]        & 2.77 & 1.87 & 2.75 & 4.52 & & 8.22 & 10.25 & 8.08 & 7.42 \\\hline
UNet-ConvLSTM-YZ  & 1.42 & 2.09 & 1.23 & 3.77 & & 4.51 & 11.30 & 3.80 & 6.17\\  
CMAViT-YZ         & 1.52 & 1.57 & 1.26 & 6.09 & & 4.54 & 8.78 & 3.86 & 9.90 \\ 
\hline\hline
CMAViT-YZ vs. CMAViT [D]&  \textcolor{teal}{+1.43} & \textcolor{teal}{+1.39} & \textcolor{teal}{+1.28} & \textcolor{teal}{+4.12} & & \textcolor{teal}{+4.50} & \textcolor{teal}{+6.97} & \textcolor{teal}{+4.10} & \textcolor{teal}{+6.80}\\
CMAViT-YZ vs. UNet-ConvLSTM-YZ &    \textcolor{red}{-0.10} & \textcolor{teal}{+0.12} & \textcolor{red}{-0.03} & \textcolor{red}{-2.32} & & \textcolor{red}{-0.03} & \textcolor{teal}{+2.52} & \textcolor{red}{-0.06} & \textcolor{red}{-3.73}\\
\hline\hline
\label{table:ch3 exresults}
\end{tabular}
\end{table*}

Furthermore, comparing the full CMAViT model (category E) with CSR to the UNet-ConvLSTM model with CSR, there is a notable improvement in MAPE for the LER, reducing from 12.37\% to 8.22\%, and for the HER, reducing from 15.13\% to 7.42\%. These results clearly indicate that the importance of additional input resources including sentinel-1, climate data and unlimited management practice information. However, the models incorporating the YZ strategy outperformed all other models. The MAPE for both UNet-ConvLSTM and CMAViT models with the YZ strategy was below 5\% for all samples, indicating high overall accuracy. Specifically, for the CMAViT model, the MAPE for the LER was 8.78\%, and for the HER it was 9.90\%. These results highlight the effectiveness of the YZ strategy in enhancing model performance, particularly in reducing errors across the entire yield distribution, including the challenging low and high extreme values. This improvement demonstrates the capability of the YZ strategy to account for spatial heterogeneity, leading to more precise and reliable yield predictions.

Overall, the findings demonstrate that additional information such as climate data, and management practices information, alongside with the CSR and YZ strategies help mitigate the bias towards the majority of samples, thereby improving the accuracy of predictions for extreme yield values. The CMAViT-based models, in particular, exhibit greater robustness in handling these challenging predictions compared to the UNet-ConvLSTM-based models, further underlining the effectiveness of incorporating these advanced techniques in agricultural yield prediction tasks.

\subsubsection{Quantitative Results of Spatio-Temporal Yield Estimation}

This section illustrates the spatio-temporal analysis of the CMAViT model (category E), applied to the MoA-04 block (shown in Figure~\ref{fig:bho}) over a 15-week period, beginning April 1st and continuing through mid-July. The analysis was conducted on a test set that was withheld from the training data distribution to ensure an unbiased evaluation of the model's performance. The results indicate a clear trend of performance improvement as more temporal data is accumulated, with the Mean Absolute Error (MAE) reducing from 1.11 t/ha in week 1 to 1.02 t/ha by week 15, and the Mean Absolute Percentage Error (MAPE) decreasing from 3.23\% to 3.09\%.

\begin{figure*}[ht]
\centering
\captionsetup{font=footnotesize}%
\includegraphics[width=16cm, height=15cm]{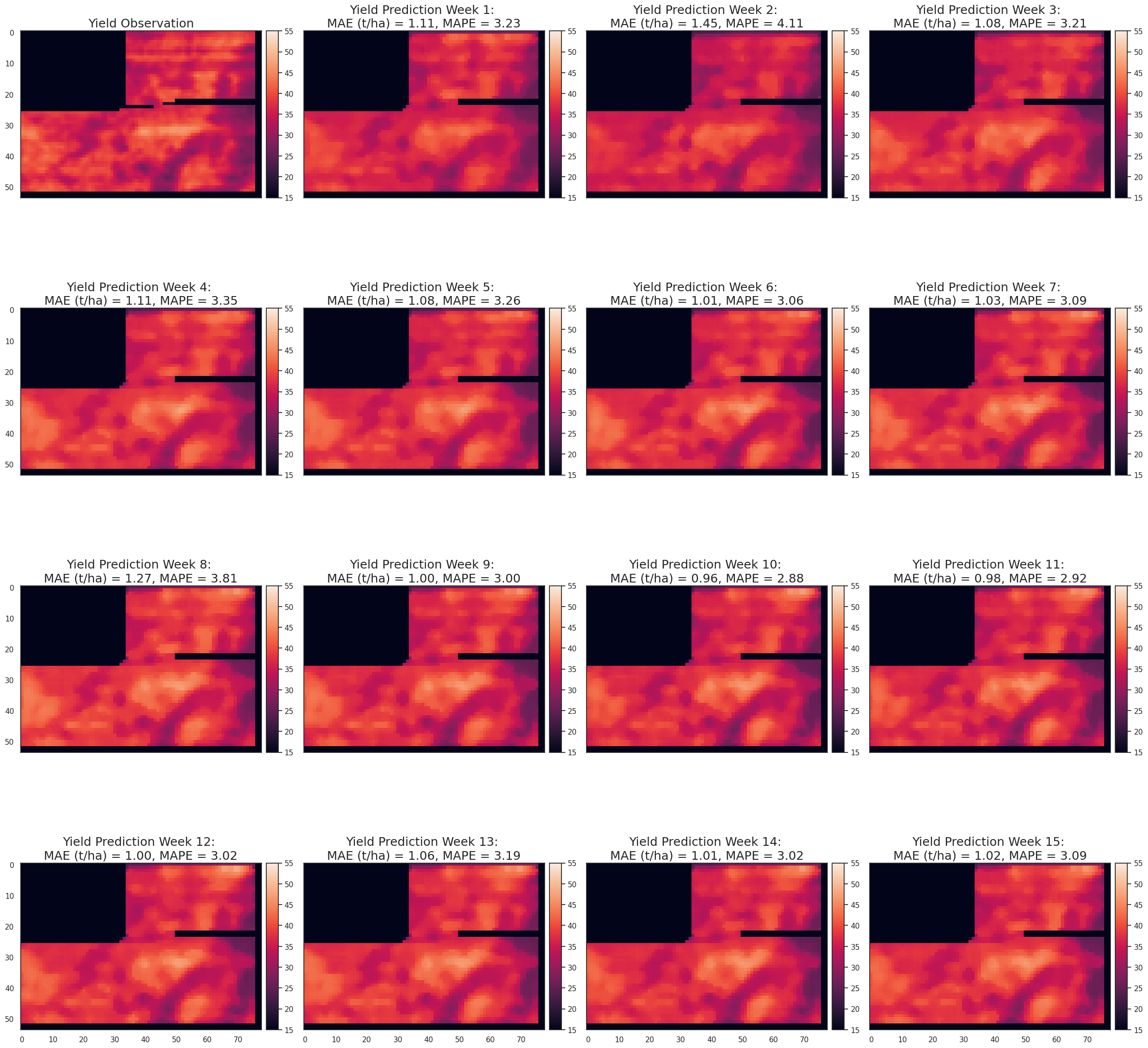} 
\caption{Spatial temporal yield prediction on MoA-04 over 15 weeks prediction. }
\label{fig:ch3 stv}
\end{figure*}

These findings underscore the benefit of accumulating information throughout the growing season, which enables the model to make more informed predictions by incorporating knowledge of crop growth dynamics over time. As the temporal series progresses, the model gains a deeper understanding of the seasonal changes and crop development stages, allowing for more accurate predictions. This highlights the value of a spatio-temporal approach in precision agriculture, where early-season estimates can be progressively refined as additional data is integrated, ultimately leading to improved prediction accuracy and better-informed management decisions. The trend observed suggests that models benefiting from richer temporal data can more effectively reduce uncertainties, thereby enhancing the reliability of yield forecasts.

\subsection{Spatial Variability Analysis}

\begin{figure*}[ht]
\centering
\captionsetup{font=footnotesize}%
\includegraphics[width=16cm, height=15cm]{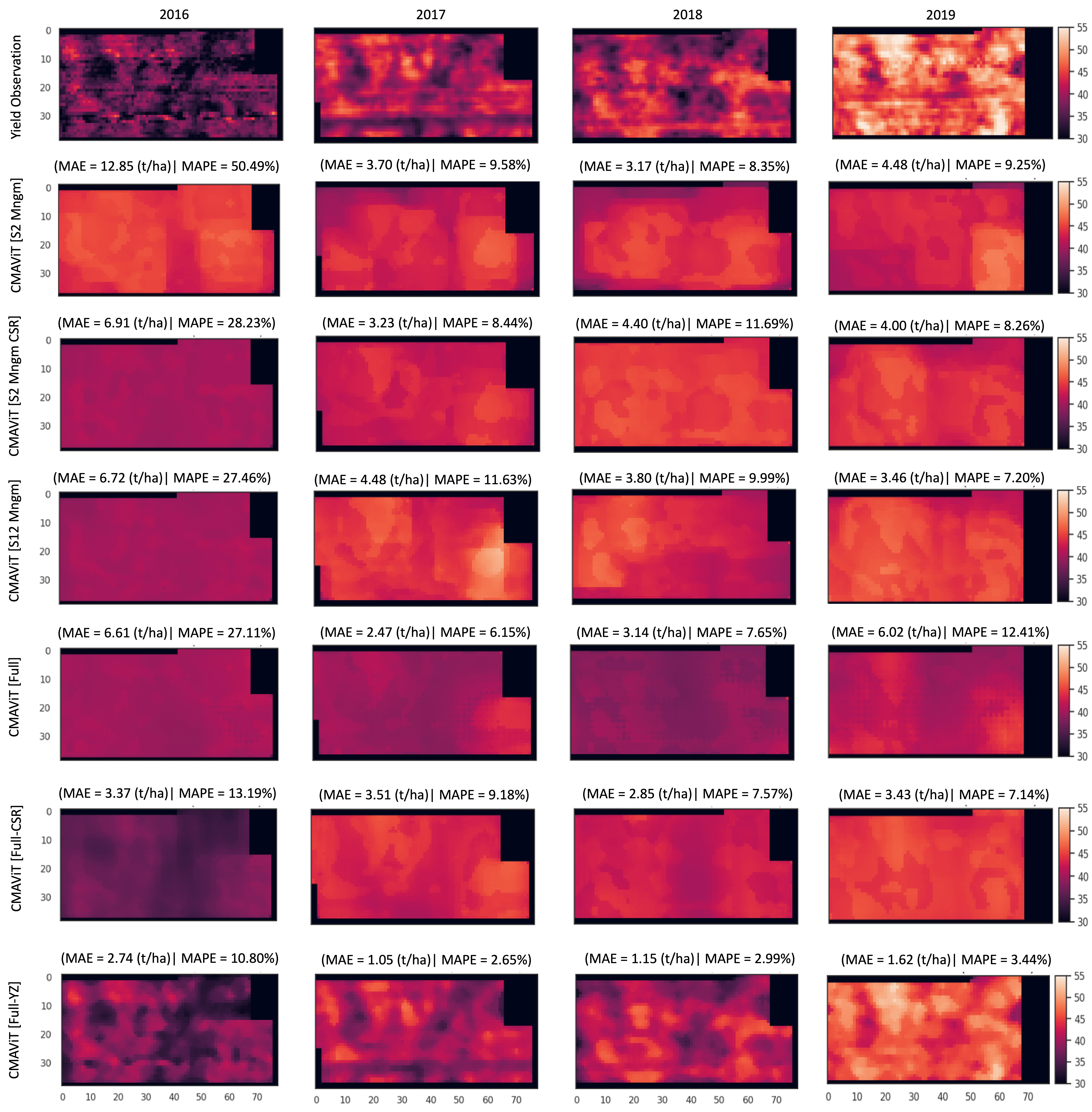} 
\caption{Evaluation of CMAViT model on Ries-06 vineyard block over 4 years observation. }
\label{fig: yearspredict}
\end{figure*}

Figure~\ref{fig: yearspredict} presents a comparative analysis of the efficacy of various CMAViT-based models in accurately predicting extreme yield values and understanding spatial variability within vineyard blocks. The evaluation was conducted on a block named Ries-06 (as yield range depicted in Figure~\ref{fig:bho}) over four years of observations. The yield in this vineyard block ranged from 10 to 65 t/ha, covering low-extreme, common, and high-extreme yield values. In 2016, the observed yields were primarily common and low-extreme, whereas 2019 showed a notable increase in high-extreme values.

The analysis reveals that the CMAViT [A] model struggled to estimate yields at the lower and higher ends of the yield spectrum, especially evident in 2016 when the yields were predominantly low-extreme. The model showed a Mean Absolute Percentage Error (MAPE) of 50.49\%, highlighting its tendency to converge toward the mean and thus demonstrating insufficient sensitivity to spatial variability within the blocks. In contrast, the CMAViT [B] model, presented in the third column, displayed a modest improvement in capturing extreme values compared to CMAViT [A], yet it continued to fall short in accurately estimating both extreme values and spatial variability.

Further improvement was observed when Sentinel-1 data (category C) was incorporated, enhancing the performance of the CMAViT model for all four years of evaluation. The results for CMAViT [D, E] of the figure present results from the CMAViT model with full input data, both with and without the CSR strategy. These models demonstrated significantly better performance in capturing both extreme yield values and spatial variability within the blocks compared to previous models, underscoring the importance of incorporating all available modalities. For instance, in 2016, the CMAViT [E] model was able to accurately capture the low-extreme and common yield values, effectively reflecting the spatial variability pattern observed in the ground truth data. This led to a substantial improvement in prediction accuracy, reducing the Mean Absolute Error (MAE) from 12.85 to 3.37 t/ha and the MAPE from 50.49\% to 13.19\%. Similarly, in 2019, a year characterized by high-extreme yield values, the model improved its performance significantly, with MAE decreasing from 4.48 to 3.43 t/ha and MAPE dropping from 9.25\% to 7.14\% compared to the CMAViT model. However, the results for the last row, which include the YieldZone (YZ) strategy, show a significant improvement in capturing spatial variability within the field. The integration of the YZ strategy led to a reduction in MAPE across all years to below 10\%, with particularly notable performance in 2017, 2018, and 2019, where the MAPE dropped to below 4\%. This demonstrates the model's enhanced capability to accurately learn and represent spatial variability, effectively capturing yield differences across different sections of the field. The incorporation of the YZ strategy not only improved the overall prediction accuracy but also allowed the model to reflect the spatial variability patterns observed within the field, resulting in a more precise depiction of the yield distribution and providing better insights for targeted management practices.

Across the years 2017 and 2018, all CMAViT-based models showed consistently superior performance compared to other years. This improvement can be attributed to the implementation of the YieldZone strategy, which enhances the model's ability to recognize similarities within input features across vineyard blocks, leading to better predictions. Moreover, the incorporation of the 'ExtremeWeight' strategy was crucial in honing the model's focus on both low and high-extreme values, thereby strengthening its precision in capturing the full spectrum of yield variability. The combination of these strategies and the inclusion of multimodal data highlights the model's capability to effectively adapt to both spatial and temporal dynamics, ultimately leading to more accurate and reliable yield predictions that can better inform vineyard management decisions.

\subsection{Modality Maskout Analysis}

Table~\ref{table:maskout} presents the results of a mask-out analysis conducted by selectively removing one of the data modalities from the CMAViT model to assess the impact of each component on the model’s performance. In the first scenario, the management practices context was masked out, resulting in a noticeable degradation in the model's performance (CMAViT [D]), with the R² value dropping from 0.80 to 0.73, and the MAPE increasing from 10.19\% to 11.92\%. This highlights the role that management practices play in predicting yield, as removing this modality reduced the model's accuracy. In the second scenario, the climate data modality was excluded. This also led to a decrease in performance, with the R² value decreasing from 0.80 to 0.70 and the MAPE increasing from 10.19\% to 12.66\%. These results emphasize the importance of incorporating climate information, which plays a crucial role in accurately modeling the yield given its influence on crop growth and environmental variability.

\begin{table*}[ht!]
\captionsetup{font=footnotesize}%
\footnotesize
\centering
\caption{Modality Maskout Performance Analysis. Metrics include R-squared (R²), Mean Absolute Error (MAE $t/ha$), Root Mean Squared Error (RMSE), and Mean Absolute Percentage Error (MAPE\%).}
\begin{tabular}{l p{0.4cm} p{0.45cm} p{0.65cm} p{1.2cm}  p{0.4cm} p{0.45cm} p{0.65cm} p{1.2cm} p{0.45cm} p{0.45cm} p{0.65cm} p{1.1cm}} 
 \toprule
    & \multicolumn{4}{c}{Train} & \multicolumn{4}{c}{Validation} & \multicolumn{4}{c}{Test} \\
    \cmidrule(r){2-5} \cmidrule(r){6-9} \cmidrule(r){10-13}

Model & $R^{2}$ & MAE & RMSE & MAPE(\%) & $R^{2}$ & MAE & RMSE & MAPE(\%) &$R^{2}$ & MAE & RMSE & MAPE(\%)\\ 
 \hline\hline
CMAViT &  0.94 & 2.14 & 2.84 & 7.34 &  0.67 & 3.82 & 5.15 & 13.87 &0.80 & 3.29 & 4.43 & 10.19\\\hline
Mngm-Maskout &  0.94& 2.27 & 2.92 & 7.66 & 0.68 & 3.79& 5.05 & 13.29 &0.73 & 3.86 & 5.08 & 11.92\\
Climate-Maskout &  0.85 & 3.55 & 4.50 & 12.48 &  0.68 & 3.81 & 5.08 & 14.10 &0.70 & 4.10 & 5.38 & 12.66\\
Mngm-Climate-Maskout  &  0.89 & 2.90 & 3.79 & 10.00 &  0.65 & 3.88 & 5.27 & 13.93 & 0.72 & 3.93 & 5.19 & 12.39 \\
\hline\hline
\label{table:maskout}
\end{tabular}
\end{table*}

Finally, in the scenario where both the management practices and climate data were removed, leaving only the Sentinel-1 and Sentinel-2 satellite data, the model's performance further declined, with the R² dropping to 0.72 and the MAPE rising to 12.39\%. This outcome shows a clear reduction in the model's ability to predict yields effectively without these critical inputs. The results from this mask-out analysis underscore the importance of each data modality. The declines in performance metrics indicate that each modality—whether management practices, climate data, or satellite imagery—contributes unique and essential information that enhances the model's capability to learn the complexities of yield prediction. The integration of all these modalities is therefore crucial for maximizing prediction accuracy, highlighting the value of a multimodal approach in agricultural modeling.

\section{CONCLUSION}
This experiment aims to develop a deep learning model to better understand the complex interactions between crop growth, climate change, and management practices using the Climate-Management Aware Vision Transformer (CMAViT) for spatio-temporal vineyard yield estimation. The CMAViT model incorporates a spatio-temporal multimodal approach to capture the dynamics between satellite imagery and climate data throughout the growing season. Additionally, it learns the interactions between time-series signals and management practices. To achieve this, we utilized a large, heterogeneous yield observation dataset spanning four years, 41 blocks, and eight grape cultivars. Based on this dataset, a spatio-temporal deep learning model was developed, which simultaneously learns both spatial and temporal correlations in yield development and provides yield forecasts for each block from April 1st to July 15th.

The results show that the CMAViT model demonstrates greater flexibility in integrating data from various sources with different structures, allowing it to better capture the complexities of vineyard yield estimation, especially when considering short-term climate variations and management practices. Compared to CNN-based models like UNet-ConvLSTM, CMAViT outperforms significantly, improving the R² score from 0.78 to 0.84 and reducing MAPE from 12.57\% to 8.22\% on unseen test data. Additionally, the YieldZone strategy further enhanced performance for extreme yield ranges, reducing MAPE by 6.97\%, 4.10\%, and 6.80\% for low, common, and high yield extremes, respectively. Overall, CMAViT's proficiency in capturing spatio-temporal relationships and integrating diverse data sources makes it a powerful tool for precise and timely vineyard yield forecasting, offering significant advantages for agricultural decision-making.

\section{Acknowledgments}
\noindent This project was partly supported by USDA AI Institute for Next
Generation Food Systems (AIFS), USDA award number 2020-67021-
32855.

\bibliographystyle{unsrtnat}

\bibliography{references}  

\end{document}